\documentclass[letterpaper, 10 pt, conference]{ieeeconf}  % Comment this line out if you need a4paper

\IEEEoverridecommandlockouts                              % This command is only needed if 
\usepackage{graphicx}
\usepackage{booktabs}
\usepackage{orcidlink}
\usepackage{multirow}
\usepackage{pifont}
\usepackage{xcolor}
\usepackage{colortbl}
\usepackage{url}
\usepackage{bbm}
\usepackage{amsmath}        % aligned equations
\usepackage{mathrsfs}
\usepackage{physics}
\usepackage{amsfonts}       % blackboard math symbols
\usepackage{wrapfig} 
\usepackage{marvosym}

\title{\LARGE \bf
Foundation Feature-Driven Online End-Effector Pose Estimation: A Marker-Free and Learning-Free Approach
}

\author{Tianshu Wu*, Jiyao Zhang*$^\dagger$, Shiqian Liang*, Zhengxiao Han, Hao Dong$^{\textrm{\Letter}}$%~\textsuperscript{\Letter}
\thanks{Authors are with Center on Frontiers of Computing Studies, School of Computer Science, Peking University, Beijing 100871, China, also with PKU-Agibot Lab, Beijing 100871, China, and also with National Key Laboratory for Multimedia Information Processing, School of Computer Science, Peking University, Beijing 100871, China.
}
\thanks{* indicates equal contribution}
\thanks{$^\dagger$ Project leader: {\tt\small jiyaozhang@stu.pku.edu.cn}}
\thanks{%\textsuperscript{\Letter} 
$^{\textrm{\Letter}}$ Corresponding to {\tt\small hao.dong@pku.edu.cn}}
}

\newcommand{\Tref}[1]{Table~\ref{#1}}
\newcommand{\Fref}[1]{Figure~\ref{#1}}

\newcommand{\Sref}[1]{Section~\ref{#1}}

\newcommand{\Yes}{\textcolor{green}{\ding{51}}}
\newcommand{\No}{\textcolor{red}{\ding{55}}}
\newcommand{\colour}{\rowcolor{gray!15}}

\newcommand{\observation}{\vb*{O}}
\newcommand{\cImg}{\vb*{I}_c}
\newcommand{\dImg}{\vb*{I}_d}
\newcommand{\Img}{\vb*{I}}
\newcommand{\state}{\vb*{s}}
\newcommand{\pose}{\vb*{p}}
\newcommand{\temNum}{K_t}
\newcommand{\refNum}{K_r}

\newcommand{\keyFrameNum}{K_f}
\newcommand{\Matches}{\mathcal{M}}
\newcommand{\singleRes}{\mathcal{S}}
\newcommand{\dino}{\Psi}
\newcommand{\dinofeat}{\mathcal{F}}
\newcommand{\uv}{\vb*{u}}
\newcommand{\pts}{\vb*{X}}

\begin{document}

\maketitle
\thispagestyle{empty}
\pagestyle{empty}

%%%%%%%%%%%%%%%%%%%%%%%%%%%%%%%%%%%%%%%%%%%%%%%%%%%%%%%%%%%%%%%%%%%%%%%%%%%%%%%%

\begin{abstract}
    Accurate transformation estimation between camera space and robot space is essential. Traditional methods using markers for hand-eye calibration require offline image collection, limiting their suitability for online self-calibration. Recent learning-based robot pose estimation methods, while advancing online calibration, struggle with cross-robot generalization and require the robot to be fully visible. 
This work proposes a \textbf{F}oundation feature-driven online \textbf{E}nd-\textbf{E}ffector \textbf{P}ose \textbf{E}stimation (\textbf{FEEPE}) algorithm, characterized by its training-free and cross end-effector generalization capabilities. Inspired by the zero-shot generalization capabilities of foundation models, FEEPE leverages pre-trained visual features to estimate 2D-3D correspondences derived from the CAD model and target image, enabling 6D pose estimation via the PnP algorithm. To resolve ambiguities from partial observations and symmetry, a multi-historical key frame enhanced pose optimization algorithm is introduced, utilizing temporal information for improved accuracy.
Compared to traditional hand-eye calibration, FEEPE enables marker-free online calibration. Unlike robot pose estimation, it generalizes across robots and end-effectors in a training-free manner. Extensive experiments demonstrate its superior flexibility, generalization, and performance. Additional demonstrations are available at \url{https://feepose.github.io/}
\end{abstract}

%%%%%%%%%%%%%%%%%%%%%%%%%%%%%%%%%%%%%%%%%%%%%%%%%%%%%%%%%%%%%%%%%%%%%%%%%%%%%%%%
\section{INTRODUCTION}
\label{sec:intro}
\begin{figure}[tbp]
\centerline{\includegraphics[width=0.48\textwidth]{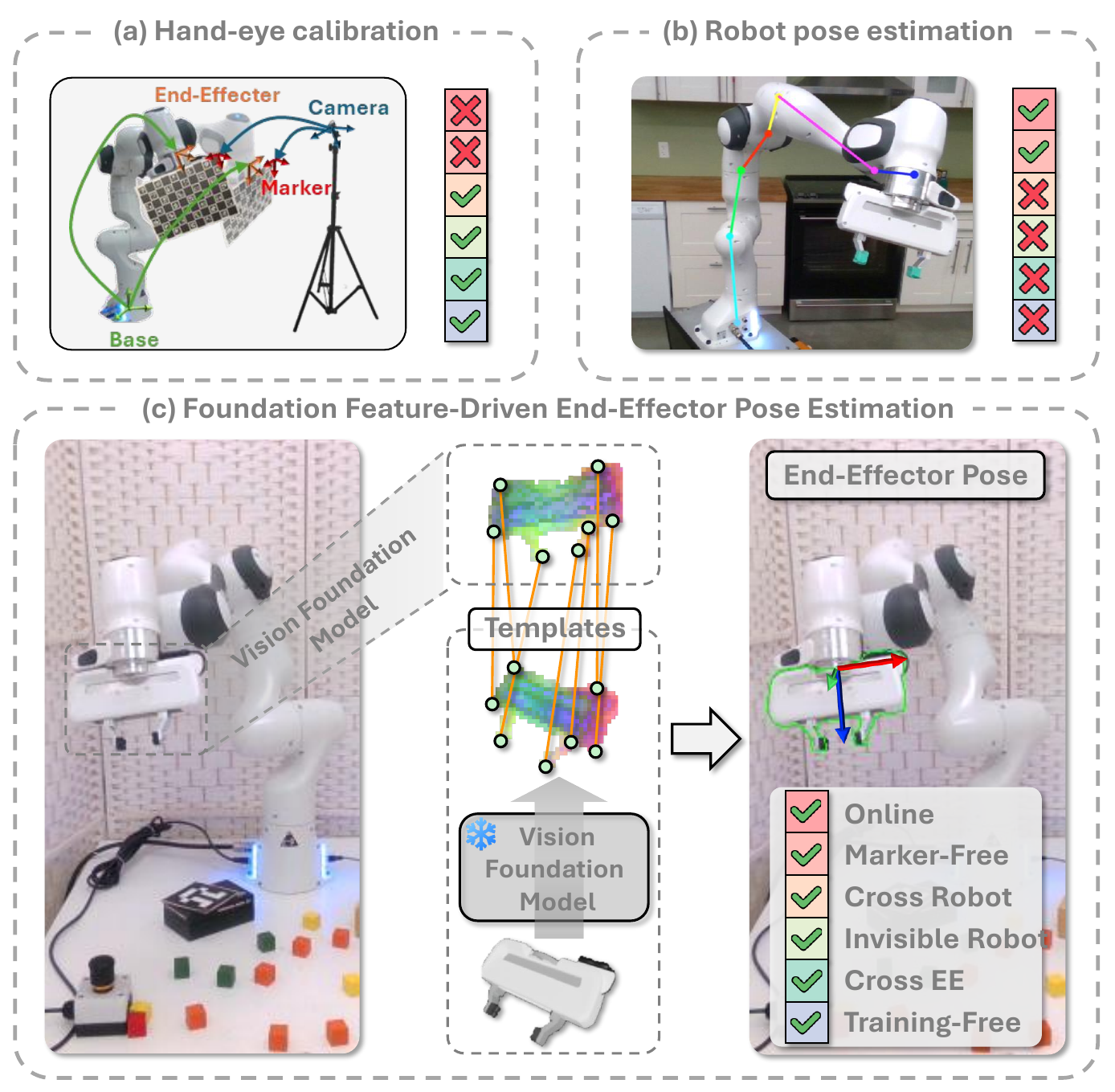}}
\vspace{-6pt}
\caption{We propose \textbf{FEEPE} (\textbf{F}oundation feature-driven \textbf{E}nd-\textbf{E}ffector \textbf{P}ose \textbf{E}stimation), an online, marker-free, training-free method for pose estimation that generalizes across robots and end-effectors.}
\vspace{-20pt}
\label{fig:teaser}
\end{figure}

Consider a robot performing a manipulation task~\cite{omnimanip, graspgf, lvdiffusor, 3dtacdex} where perception results are obtained in the camera space. How can a robot execute actions based on these perception results? This requires an accurate transformation between the camera and the robot space. Traditional methods employ augmented reality (AR) tags~\cite{garrido2014automatic,fiala2005artag,olson2011apriltag} as markers attached to the end-effector and solve a homogeneous matrix equation~\cite{fassi2005hand} to determine the transformation. However, this approach requires an offline collection of images of the robotic arm in different states for optimization, making it unsuitable for online robot self-calibration~\cite{du2013online,du2015online}. This limitation restricts the rapid deployment of robotic systems.

Recent advancements in learning-based robot pose estimation algorithms~\cite{lee2020camera,lu2022pose,labbe2021single} have shown promise for enabling online self-calibration. These algorithms~\cite{lu2023markerless,tian2024robokeygen} aim to use data-driven methods to estimate a robot's pose from images.
However, these methods lack cross-robot generalization and require full robot visibility, limiting their applicability.
Another related field is object pose estimation~\cite{genpose, omni6dpose, wen2023foundationpose}, which could be adapted for end-effector pose estimation. However, these methods typically require training on large-scale object pose estimation datasets and struggle with symmetric objects, which is common in end-effectors.
Our goal is to develop an \textbf{online}, \textbf{marker-free}, \textbf{highly generalizable} and \textbf{training-free} end-effector pose estimation algorithm for robot self-calibration, which presents the following challenges:

\begin{itemize}
    \item End-effector diversity in appearance and geometry poses a challenge for cross-end-effector generalization.
    \item The ambiguity issue caused by partial observations and end-effector symmetry.
\end{itemize}

To address these, we propose \textbf{FEEPE} (\textbf{F}oundation feature-driven \textbf{E}nd-\textbf{E}ffector \textbf{P}ose \textbf{E}stimation). 
% Inspired by the zero-shot generalization ability~\cite{zhang2024tale,amir2021deep} of visual foundation models~\cite{kirillov2023segment} like DINOv2~\cite{oquab2023dinov2}, the algorithm leverages pre-trained visual features for end-effector 6D pose estimation, effectively handling the appearances and geometry variation.
Leveraging the zero-shot generalization of visual foundation models~\cite{kirillov2023segment,oquab2023dinov2}, FEEPE uses pre-trained features for 6D pose estimation, handling appearance and geometry variations.
% Specifically, with the CAD model of the end-effector known, multiple reference images from different perspectives are pre-rendered first. The DINOv2 algorithm then extracts features from these images and the target image to establish 2D-2D correspondences. These correspondences facilitate the creation of 2D-3D mappings, linking the target image to the 3D model.
Given a CAD model, we pre-render reference images. We use Dinov2 to extract features and establish 2D-2D correspondences, enabling 2D-3D mappings to the 3D model.
This facilitates an initial 6D pose estimation using the PnP algorithm~\cite{lepetit2009ep}. Nevertheless, the ambiguity from partial observations and the end-effector's symmetry can lead to inaccuracies when predicting from a single image. To address this, we introduce a multi-historical key frame enhanced pose optimization that utilizes temporal information and robot priors to resolve symmetry ambiguities and enhance accuracy.

Overall, as shown in ~\Fref{fig:teaser}, FEEPE enables marker-free online calibration compared to traditional hand-eye calibration. In contrast to learning-based robot pose estimation, our method achieves cross-robot and cross-end-effector generalization in a training-free manner and does not require the robot arm to be visible.
This is the first online, marker-free, generalizable, training-free end-effector pose estimation algorithm for robot self-calibration.
Extensive experimental validation demonstrates the convenience, robustness, and high precision (1mm) of our approach, whether compared to learning-based or traditional hand-eye calibration methods.

%%%%%%%%%%%%%%%%%%%%%%%%%%%%%%%%%%%%%%%%%%%%%%%%%%%%%%%%%%%%%%%%%%%%%%%%%%%%%%%%

\section{RELATED WORK}
\label{sec:related}
% We aim to solve for the \textbf{camera-to-robot pose} using the \textbf{CAD model} of the end-effector, hence relating to works on \textbf{camera-to-robot pose estimation} and \textbf{CAD model-based object pose estimation}.

\subsection{Camera-to-Robot Pose Estimation.}
Traditionally, camera-to-robot pose is estimated by hand-eye calibration~\cite{fassi2005hand}, using markers like ARTag~\cite{fiala2005artag} or AprilTag~\cite{olson2011apriltag}. Recently, learning-based methods~\cite{lu2022pose,tian2023robot} have emerged, utilizing deep neural networks for online calibration. DREAM~\cite{lee2020camera} leverages a CNN to detect pre-defined robot keypoints and solve the camera-to-robot pose using a PnP solver. CtRNet~\cite{lu2023markerless} further improves the performance by utilizing a differentiable renderer and a segmentation objective. RoboKeyGen~\cite{tian2024robokeygen} uses a diffusion model to lift 2D keypoints into 3D, jointly estimating robot joint angles and camera pose. RoboPose~\cite{lu2022pose} adopts a render-and-compare approach. Current methods have taken a step towards online calibration, but they require the full robot to be visible and are robot-specific, whereas our method generalizes to unseen robots.
There have also been works aiming to perform online calibration by end-effector pose estimation. ~\cite{cheng2020real} directly regresses the end-effector pose from a pointcloud. ~\cite{sefercik2023learning} predict 3D positions of the end-effector and solve for the effector pose, then calibrate using multiple estimations. Such methods are all end-effector specific, while our method can generalize to unseen end-effectors without training.

\subsection{CAD Model-based Object Pose Estimation.}
Early methods of object pose estimation~\cite{xiang2017posecnn,park2019pix2pose} estimate the 6D pose of a known object. These methods are instance-level~\cite{wang2019densefusion,li2019cdpn}, meaning that the test object is seen in training~\cite{he2020pvn3d,he2021ffb6d,labbe2020cosypose}, and the method is unable to generalize to unseen objects. To relax this constraint, recent efforts aim to estimate the unseen object pose with the textured CAD model known. ~\cite{gou2022unseen} proposed the challenge of novel object pose estimation, providing a method that established correspondences between the object pointcloud and the scene pointcloud. ~\cite{hagelskjaerkeymatchnet, lin2023sam,huang2024matchu} follow this path, seeking to directly extract object-agnostic features~\cite{zhao2023learning} from the CAD model and match them with features of the scene to obtain 3D-3D matches. Other works~\cite{labbe2022megapose,wen2023foundationpose,moon2024genflow} use a render-and-compare approach, iteratively refining a coarse estimate by rendering the object in different poses and comparing it with the target image. Template-based methods~\cite{shugurov2022osop,ausserlechner2023zs6d} render templates~\cite{nguyen2023gigapose,wang2024object}, and retrieve the nearest template during test time~\cite{ornek2023foundpose,nguyen2022templates}, and perform further refinement or optimization. While progressing towards estimating the unseen object pose, all current approaches struggle with ambiguity caused by partial observations and symmetry. Our approach enhances pose estimation by addressing symmetry ambiguities and improving accuracy through the integration of temporal information and robot priors.

%%%%%%%%%%%%%%%%%%%%%%%%%%%%%%%%%%%%%%%%%%%%%%%%%%%%%%%%%%%%%%%%%%%%%%%%%%%%%%%%

\section{METHOD}
\label{sec:method}
\noindent\textbf{Problem Formulation.}
We assume the 3D model of end-effector known and take a sequence of observations \(\{\observation_i\}_{i=1}^t\), where \(\observation = \{\cImg, \dImg, \state\}\). Here, \(\cImg\) represents the RGB image, \(\dImg\) represents the depth image, and \(\state\) represents the states of the robotic arm. The goal is to predict the pose of the end-effector at time \(t\), denoted as \(\pose_t \in {SE(3)}\).

\begin{figure*}[tbp]
\vspace{5pt}
\centerline{\includegraphics[width=0.99\textwidth]{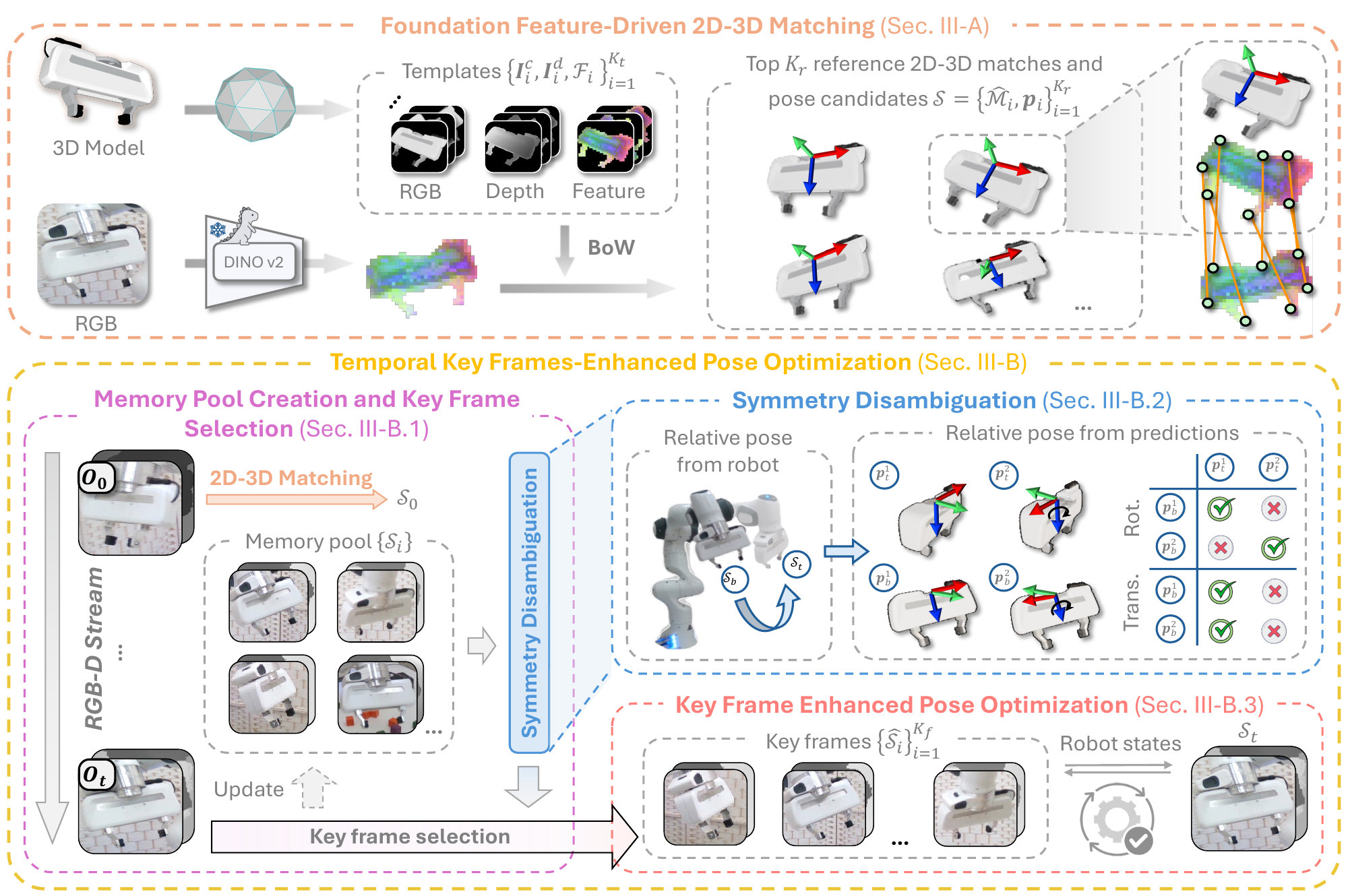}}
\vspace{-10pt}
\caption{\textbf{Overview of FEEPE.} Given the 3D model of the end-effector and a target image, we first render multi-view templates. Using foundation features, we find the top $\refNum$ references most similar to the target image and compute 2D-3D matches and pose candidates (\Sref{sec:2d-3d_matching}). To address ambiguities from partial observations, we introduce a global memory pool (\Sref{sec:memory_pool}) that records keyframes and robot states for pose optimization (\Sref{sec:pose_optimization}). To resolve ambiguities from symmetry, we propose a symmetry disambiguation module (\Sref{sec:symmetry_disambiguation}) to eliminate incorrect matches.
}
\vspace{-18pt}
\label{fig: method}
\end{figure*}

\noindent\textbf{Overview.} 
~\Fref{fig: method} illustrates our pipeline. With the 3D model known, we first use rendered images as references and employ foundation features to establish 2D-3D matches between the reference and target images for pose estimation, as detailed in~\Sref{sec:2d-3d_matching}. Then, to address ambiguities arising from partial observations and end-effector symmetry, we incorporated a multi-historical key frame enhanced pose optimization algorithm, as detailed in~\Sref{sec:temporal_opt}.

\subsection{Foundation Feature-Driven 2D-3D Matching}
\label{sec:2d-3d_matching}

This section elaborates on the procedures for generating reference views, establishing 2D-3D correspondences, and estimating initial pose candidates. We first render \(\temNum\) template images from the 3D model of the end-effector. We then extract foundation features using Dinov2~\cite{oquab2023dinov2} and identify the top \(\refNum\) reference views that exhibit the highest similarity to the target image. 
Based on these features, we establish 2D-3D matches to compute pose candidates with the Perspective-n-Point (PnP) algorithm~\cite{lepetit2009ep}.
In this manner, for every target image, we obtain \( \singleRes = \{ \hat{\Matches_i}, {\pose_i} \}_{i=1}^{\refNum} \), where \( \hat{\Matches_i} \) represents the inlier 2D-3D matches and \( \pose_i \in SE(3) \) denotes the pose candidates. The index \( i \in \{0, 1, ..., \refNum\} \) represents each reference view. In the following, we will delineate the specific details.

\subsubsection{Templates generation and reference views selection} 
We employ Fibonacci Sphere~\cite{gonzalez2010measurement} to sample 80 positions on a unit sphere. For each viewpoint, we uniformly sample 12 in-plane rotations, resulting in 960 sampled viewpoints. Using Blender, we render the templates with ray tracing and extract pixel-level visual features using Dinov2-ViTL~\cite{oquab2023dinov2}, denoted as \( \dino \). The visual features \( \dinofeat \) for each template image are computed as \( \dinofeat = \dino(\cImg) \), where \( \cImg \) represents the RGB images. This process results in the template set \(\{\Img_i^c, \Img_i^d, \dinofeat_i\}_{i=1}^{\temNum}\), where \(\Img_i^c\) represents the RGB images, \(\Img_i^d\) represents the depth images, \(\dinofeat_i\) represents the visual features and $\temNum=960$. Further, we construct Bag-of-Words descriptors (BoW) using the template features \(\{\dinofeat_i\}_{i=1}^{\temNum}\), following~\cite{ornek2023foundpose}. Based on these descriptors, we select the top \(\refNum\) views that are most similar to the target image as reference views. In this way, for each target image, we can identify \(\refNum\) reference images for subsequent calculations. In this paper, we set the number of reference images \(\refNum\) to 5.

\subsubsection{2D-3D matching and pose candidate estimation} 
Although we identified reference views from the templates that closely match the target image using Bag-of-Words descriptors, the discrete sampling within the templates still results in significant pose errors between the reference views and the target image. To address this issue, we utilize foundation features to establish pixel-level correspondences between the reference views and the target image. In this way, for each reference view, we can further establish 2D-3D matches between the target image and the 3D model, denoted as \(\Matches = \{\uv \leftrightarrow \pts\}\), where \(\uv\) and \(\pts\) represent 2D image points and 3D model points, respectively. Specifically, we first compute 2D-2D matches by the cosine similarity between \(\dinofeat_{ref}\) and \(\dinofeat_{tar}\), where \(\dinofeat_{ref}\) indicates the pixel-level features of the reference views and \(\dinofeat_{tar}\) indicates the pixel-level features of the target image. By incorporating the rendered depth maps, for each reference view, we convert these 2D-2D matches into 2D-3D matches $\Matches$ and recover the pose candidates $\pose$ with PnP. To remove mismatches in $\Matches$, we eliminate outliers in the PnP solving process, resulting in refined matches $\hat{\Matches}$. Through this way, for each target frame, we obtain the results \(\singleRes = \{\hat{\Matches_i}, \pose_i\}_{i=1}^{\refNum}\).
\subsection{Multi Historical Key Frame Enhanced Pose Optimization}
\label{sec:temporal_opt}
In the preceding discussion, we established the formation of 2D-3D matches and corresponding pose candidates $\singleRes = \{\Matches_i, \pose_i\}_{i=1}^{\refNum}$ between target frame and reference images. However, estimating pose from a single frame introduces ambiguities due to partial observations and symmetry. To address these, this section introduces the temporal data and robotic priors enhanced pose optimization. Specifically, ~\Sref{sec:memory_pool} discuss the memory pool creation and key frame selection, ~\Sref{sec:symmetry_disambiguation} introduce the symmetry disambiguation with temporal information and robotic priors, and ~\Sref{sec:pose_optimization} propose a key frame-enhanced joint pose optimization approach for accurate pose estimation.

\subsubsection{Memory Pool Creation and Key Frame Selection}
\label{sec:memory_pool}

To enhance the accuracy and robustness of pose estimation, we maintain a global memory pool $\{\singleRes_i\}$, where $\mathcal{S} = \{\hat{\Matches_i}, \pose_i\}_{i=1}^{\refNum}$ as described in~\Sref{sec:2d-3d_matching}. The memory pool is used to store historical data for subsequent optimization~(\Sref{sec:pose_optimization}). 
To ensure that only significantly different poses are retained, we define a criterion based on the angular distance. The angular distance function $\Omega$ is defined as follows:
\begin{equation}
\Omega(\pose_1, \pose_2) = \arccos\left(\frac{tr(\pose_1^T \pose_2) - 1}{2}\right)
\end{equation}
where $\pose_1, \pose_2 \in SO(3)$. We implement the updating criterion for the memory pool following~\cite{wen2021bundletrack}: a new frame $\observation_t$ is added only if the angular distance between the estimated $\pose_t$ and the closest existing poses $\{\pose\}$ in the memory pool satisfies: 
\begin{equation}
\min_{\pose_j \in \{\pose\}} \Omega(\pose_j, \pose_t) > \theta
\end{equation}
where $\theta$ indicates a predefined threshold, which set to $10^\circ$ in this paper. To balance computational efficiency and estimation accuracy in pose estimation, we select key frames from the memory pool for subsequent optimization. Initially, we estimate the pose \(\hat{\pose}_t\) for the new frame $\observation_t$ using the end-effector pose \(\pose_{t-1}\) of $\observation_{t-1}$ and the transformation \(\delta_{\pose}\) derived from the forward kinematics of robot:

\vspace{-10pt}
\begin{equation}
\hat{\pose}_t = \pose_{t-1} \cdot \delta_{\pose}
\end{equation}
\vspace{-15pt}

We utilize the Farthest Point Sampling (FPS)~\cite{eldar1997farthest} method to select key frames from the memory pool, starting from the estimated pose $\hat{\pose}_t$. This method aims to effectively cover the pose space by maximizing the angular distances between selected frames.
The objective for selecting key frames is succinctly formulated as:
\vspace{-2pt}
\begin{equation}
\mathcal{P}^* = \mathop{\arg\max}_{\mathcal{P} \subset \mathcal{P}_m, |\mathcal{P}| = \keyFrameNum} \min_{\pose_i, \pose_j \in \mathcal{P}, i \neq j} \Omega(\pose_i, \pose_j)
\end{equation}
\vspace{-2pt}

where \(\mathcal{P}_m\) is the set of poses corresponding to each frame in the memory pool \( \{\singleRes\} \), and \(\mathcal{P}\) is the subset of chosen from \(\mathcal{P}_m\), with \(\keyFrameNum\) being the desired number of frames. This ensures the diversity of selected frames, thereby enhancing the robustness and efficiency of the pose estimation process.

\subsubsection{Symmetry Disambiguation}
\label{sec:symmetry_disambiguation}
While the memory pool provides valuable historical data for pose estimation, the presence of symmetric end-effectors poses significant challenges. Incorrect symmetric predictions can greatly affect optimization processes. To mitigate these issues, we introduce a \textit{symmetry disambiguation} module with temporal information and robot priors. Specifically, we select a base frame $\singleRes_b$ from the memory pool, which exhibits a bimodal distribution due to symmetry, as $\{\pose_b^1, \pose_b^2\}$. For the current frame $\singleRes_t$, denote as $\{\pose_t^1, \pose_t^2\}$. The optimal pose combination follows:

\begin{equation}
\pose_b^*, \pose_t^* = \mathop{\arg\min}_{\pose_b \in \{\pose_b^1, \pose_b^2\}, \pose_t \in \{\pose_t^1, \pose_t^2\}} \Omega(\delta\pose, \delta\pose')
\end{equation}
\vspace{-2pt}
where $\delta\pose$ denotes the relative pose from forward kinematics, and $\delta\pose' = \pose_b^T \cdot \pose_t$ denotes the relative pose from prediction.

\subsubsection{Key Frame Enhanced Pose Optimization}
\label{sec:pose_optimization}
Upon obtaining key frames, the relative pose between these frames and the current frame is established with forward kinematics, defined as $\{\delta\pose_i\}_{i=1}^{\keyFrameNum}$, for optimizing $\pose_t$.
As a result, we have a set of observations, $\{\observation_i\}_{i=1}^{\keyFrameNum}$, where $\observation = \{\cImg, \dImg, \Matches, \delta\pose\}$, where $\Matches = \{\uv \leftrightarrow \pts\}$. 
For optimization, the data from all reference views in each key frame with the correct symmetry estimation are applied. 
The optimization objectives include two main parts. First, we aim to minimize the 2D reprojection error. The loss function is defined as:
\vspace{-2pt}
\begin{equation}
    \mathcal{L}_{2D} = \sum_{i=1}^{\keyFrameNum}\sum_{j=1}^{\refNum}\rho\left( \| \pi(\pose_t \cdot \delta\pose_{ij} \cdot \pts_{ij}) - \uv_{ij} \|_2 \right)
\end{equation}
\vspace{-2pt}
where $\pi(\cdot)$ represents the projection function, and $\rho(\cdot)$ is a robust cost function, Caushy loss, which mitigates the influence of outliers.
Second, we aim to minimize the 3D distance error, and the loss function is defined as:
\vspace{-2pt}
\begin{equation}
    \mathcal{L}_{3D} = \sum_{i=1}^{\keyFrameNum}\sum_{j=1}^{\refNum}\rho\left( \| \pi^{-1}(\uv_{ij}) - \pose_t \cdot \delta\pose_{ij} \cdot \pts_{ij} \|_2 \right)
\end{equation}
\vspace{-2pt}
where $\pi^{-1}(\cdot)$ serves as the back-projection function that transforms 2D image coordinates $\uv$ into 3D spatial coordinates $\pts$ with the depth known.
Combining the two items, the complete loss function is formulated as:
\vspace{-2pt}
\begin{equation}
    \mathcal{L} = \mathcal{L}_{2D} + \lambda \mathcal{L}_{3D}
\end{equation}
\vspace{-2pt}
where $\lambda$ is weighting factors, which is set to 1 in this paper.

%%%%%%%%%%%%%%%%%%%%%%%%%%%%%%%%%%%%%%%%%%%%%%%%%%%%%%%%%%%%%%%%%%%%%%%%%%%%%%%%

\section{EXPERIMENTAL RESULTS}
\label{sec:experiment}
\begin{table*}[h!]
\vspace{10pt}
\centering
\begingroup
\setlength\tabcolsep{3pt}
\caption{
\textbf{Quantitative comparison with model-based object pose estimation methods.} Values represent the AUC of ADD and ADD-S metrics with a threshold of 1 cm, where values to the left of `/' are ADD metrics and values to the right are ADD-S metrics. MegaPose corresponds to MegaPose-RGBD~\cite{labbe2022megapose}, and MegaPose† represents MegaPose-RGB~\cite{labbe2022megapose} + multi-hypothesis + ICP.
}

\begin{tabular}{c|c|c|c|c|c|c|c|c|c}
\toprule
Method & Franka Panda & Kinova-3F   & Shadow Hand & Robotiq-3F  & Robotiq-2F85 & Robotiq-2F140 & RealSense-Feanka & Average & speed(ms)\\ \midrule \midrule
MegaPose     & 59.21/68.10  & 20.02/26.85 & 2.89/9.29  & 29.66/37.98 & 22.79/51.80  & 22.12/52.29   & 0.19/1.03  & 22.41/35.33 & 1230   \\
MegaPose\dag & 65.09/70.63  & 46.29/50.58 & 45.01/46.03 & 63.62/66.88 & 23.42/56.63  & 42.85/64.05   & 13.88/38.11 & 42.88/56.13 & 2980 \\
SAM6D             & 78.22/80.90  & 80.41/81.91 & 71.77/74.34 & 79.97/86.26 & 38.79/77.27  & 39.35/75.84   & 27.48/47.68 & 59.43/74.89 & \textbf{57} \\
FoundationPose    & 81.48/83.79  & \textbf{87.13}/\textbf{87.69} & 59.51/65.99 & 82.83/87.85 & 41.12/79.41  & 35.96/69.54   & 19.94/47.17 & 58.28/74.49 & 1760\\
\colour \textbf{FEEPE(Ours)}  & \textbf{85.24}/\textbf{86.29}  & 85.24/86.36 & \textbf{84.53}/\textbf{85.23} & \textbf{88.73}/\textbf{91.67} & \textbf{84.33}/\textbf{86.11}  & \textbf{84.87}/\textbf{86.15}   & \textbf{29.27}/\textbf{48.71}  & \textbf{77.46}/\textbf{81.50} & 67\\ 
\bottomrule
\end{tabular}

\vspace{-10pt}
\label{table:baseline_ins_pose}
\endgroup
\end{table*}

In this section, we compare our method with CAD model-based object pose estimation methods and camera-to-robot pose estimation methods. Additionally, we conducted a quantitative comparison with traditional Marker-based hand-eye calibration, showcasing the high precision of our approach. 

\subsection{Experiment Setup}
\noindent\textbf{Dataset.} 
We consider two datasets: RealSense-Franka~\cite{tian2024robokeygen} and our synthetic dataset SynEEPose.
RealSense-Franka includes 4 video sequences of a Franka Panda, from which we extracted segments of 565, 500, 451, and 480 frames.
SynEEPose is a synthetic dataset generated with Blender by ray-tracing~\cite{DREDS}. We selected six commonly used end-effectors and three robot arms, combining them to form various robot configurations. Each configuration was rendered in 10 video segments, each containing 300 frames. 
In total, SynEEPose comprises 180 video segments, amounting to 54,000 frames. The selected robot arms are Franka Panda, UR10e, and UR5e, while the end-effectors include Franka Panda, Robotiq-2F85, Robotiq-2F140, Kinova-3F, Robotiq-3F, and Shadow Hand. These combinations cover a range of features such as weak textures, symmetric designs, parallel and three-finger grippers, and dexterous hands. During the generation of SynEEPose, we applied domain randomization to the background, ambient lighting, camera poses, and the states of the robot arms. 
Please refer to our website for a visualization of data samples.

\noindent\noindent\textbf{Metric.} We follow instance-level pose estimation, using the Average Distance (ADD) and ADD-S~\cite{Xiang_Schmidt_Narayanan_Fox_2018} as metrics.

\subsection{Comparison with CAD Model-based Pose Estimation}
We compare our method with FounationPose~\cite{wen2023foundationpose}, MegaPose~\cite{labbe2022megapose}, and SAM-6D~\cite{lin2023sam}, the CAD model-based pose estimation methods. To ensure a fair comparison, we provide all methods with ground truth segmentation masks.

\begin{figure}[tbp]
\centerline{\includegraphics[width=0.48\textwidth]{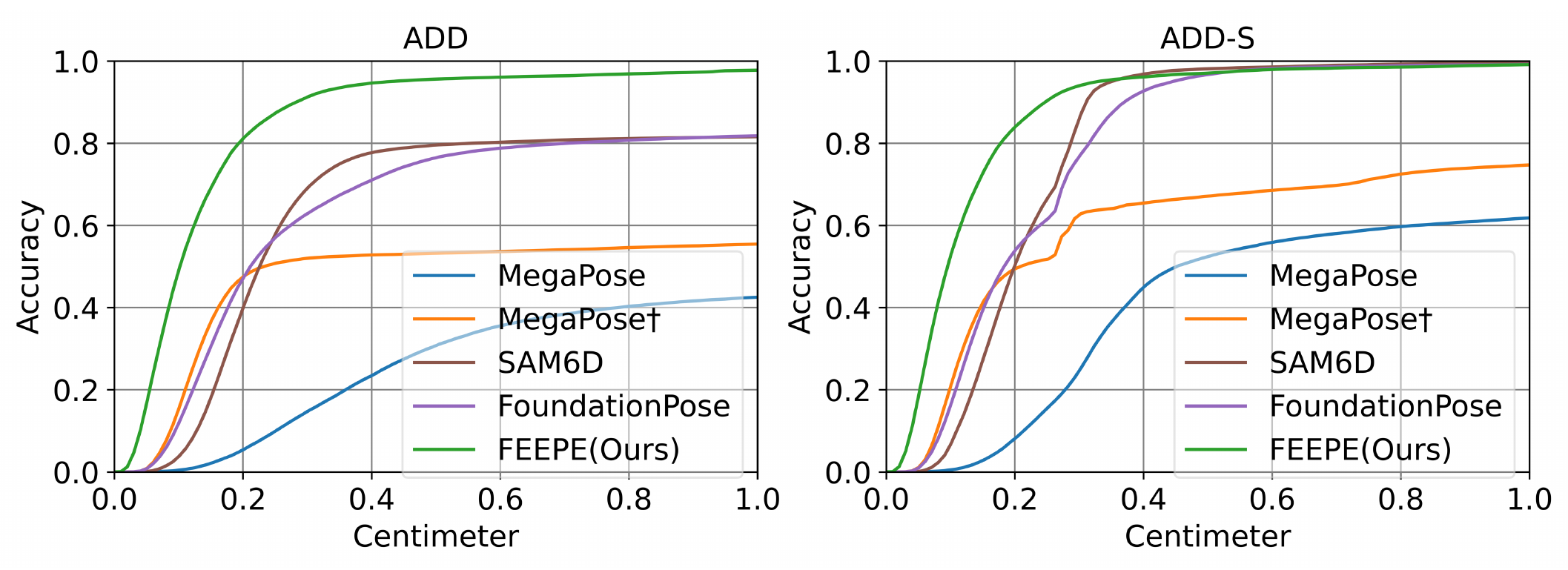}}
\caption{\textbf{Averages accuracy curves of different methods.} MegaPose† represents MegaPose-RGB~\cite{labbe2022megapose} with multi-hypothesis and ICP, and MegaPose corresponds to MegaPose-RGBD~\cite{labbe2022megapose}. 
}
\label{fig:quatitative_curve}
\end{figure}

\noindent\textbf{Accuracy analysis.}
As shown in~\Fref{fig:quatitative_curve}, our method consistently outperforms others, especially in the low-distance threshold region requiring high accuracy.
This superior performance is further validated by the quantitative results in~\Tref{table:baseline_ins_pose}, which reports the AUC values for ADD and ADD-S metrics with a threshold of 1 cm. 
Notably, while other methods have been trained on large 6D pose estimation datasets, our method achieves superior performance without training on such datasets. The results highlight the robustness and effectiveness of our approach, particularly in scenarios involving symmetrical end-effectors. 
It should be clarified that the performance difference between RealSense-Franka and SynEEPose is due to RealSense-Franka being a manually annotated dataset for robot pose estimation with errors in end-effector pose annotations.

\noindent\textbf{Run-time analysis.}
As shown in \Tref{table:baseline_ins_pose}, we report the speeds of pose estimation for different algorithms. It is important to note that these times do not include the segmentation stage. Our method demonstrates the capability for real-time online pose estimation of the end-effector.

\subsection{Ablation Study}
\begin{wraptable}{r}{0.21\textwidth}
\centering
\vspace{-16pt}
\caption{\textbf{Ablation study of critical design choices.}}
\resizebox{0.21\textwidth}{!}{\begin{tabular}{ccc}
\toprule
Method     &  ADD  & ADD-S\\ \midrule \midrule
A   & 7.50  & 19.49 \\
%dino+pnp+reproj     & 11.74 & 24.57 \\
B  & 54.28 & 70.67 \\
C & 74.40 & 79.04 \\
\textbf{D}  & \textbf{77.46} & \textbf{81.50} \\ 
\bottomrule
\end{tabular}}
\label{table: different_modules}
\end{wraptable}

\noindent\textbf{Ablation study of critical design choices.}
We test four different setups of our method: \textbf{A:} Only using Dinov2 2D-3D matching and solving for the end-effector pose through PnP; \textbf{B:} Performing optimization with only one frame on top of A; \textbf{C:} Performing temporal key frames-enhanced pose optimization on top of A without symmetry discrimination; \textbf{D:} Our full method, adding symmetry discrimination on top of C. 
Results are shown in \Tref{table: different_modules}. Notably, method B achieves comparable to SAM6D~\cite{lin2023sam} and FoundationPose~\cite{wen2023foundationpose} by only using single frame observations. Results of methods C and D also show that multi-frame optimization and symmetry discrimination both boost our performance significantly, validating the effectiveness of these designs.

\begin{figure}[h]
\centering
\includegraphics[width=0.46\textwidth]{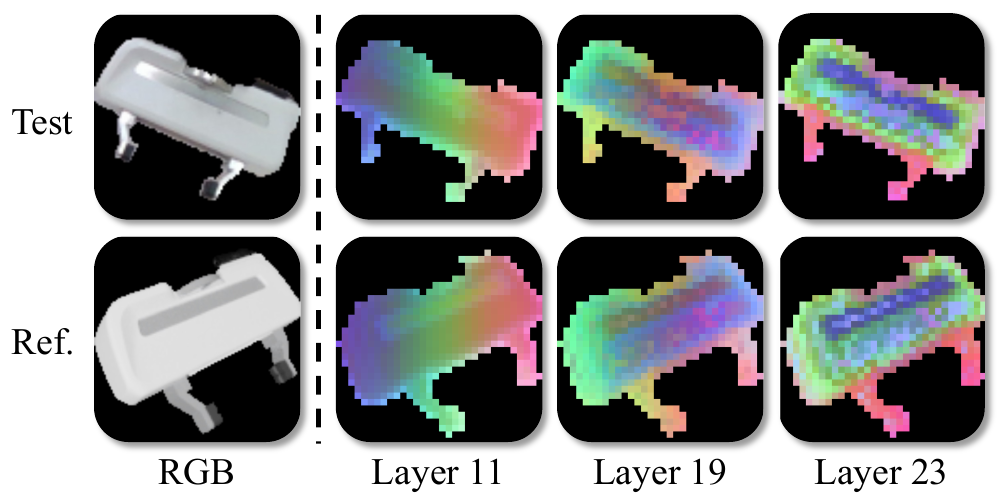}
\vspace{-10pt}
\caption{\textbf{Features visualization from various Dinov2 layers.}}
\vspace{-12pt}
\label{fig:layers}
\end{figure}

\begin{wraptable}{r}{0.21\textwidth}
\vspace{-6pt}
\centering
\caption{\textbf{Effects of different Dinov2 layers.}}
\resizebox{0.21\textwidth}{!}{% \begin{table}[]
\begin{tabular}{cccc}
\toprule
Layer & ADD & ADD-S \\ \midrule \midrule
11 & 73.05 & 77.84 \\
\textbf{19} & \textbf{77.46} & \textbf{81.50} \\
23 & 67.86 & 73.60 \\
\bottomrule
\end{tabular}
% \end{table}
}
\label{table: different_layers}
\end{wraptable}
\noindent\textbf{Effects of the Layers of Dinov2.} We visualize the features extracted at different layers of Dinov2 in~\Fref{fig:layers}. 
With deeper layers, semantic features improve while positional features weaken~\cite{amir2021deep}, affecting matching accuracy, especially for symmetric end-effectors. ~\Tref{table: different_layers} shows results using feature descriptors from different layers, which offer an effective trade-off at intermediate layers.

\begin{table*}[h]
\vspace{5pt}
\centering
\caption{\textbf{Effects of hyperparameters.} 
\textbf{Bold} numbers indicate hyperparameters selected in the end.}
\begin{tabular}{c|cccc|cccc|cccc}
\toprule
Hyperparameters & \multicolumn{4}{c|}{\textbf{Templates}} & \multicolumn{4}{c|}{\textbf{Reference views}} & \multicolumn{4}{c}{\textbf{Keyframes}} \\ \cmidrule{2-13}
 & 240 & 480 & \textbf{960} & 1920 & 1 & 3 & \textbf{5} & 10 & 2 & 4 & \textbf{8} & 16 \\
\midrule \midrule
ADD  & 73.08  & 75.06 & 77.46 & 77.66 & 64.21 & 75.40  & 77.46 & 77.98 & 73.75 & 76.26 & 77.46 & 77.72  \\
ADD-S  & 77.82  & 79.53 & 81.50 & 81.64 & 69.10 & 79.57 & 81.50 & 81.86 & 78.54 & 80.60 & 81.50 & 81.71 \\
Speed(ms)  & 63 & 65 & 67 & 71 & 55 & 61 & 67 & 81 & 54 & 59 & 67 & 86 \\
\bottomrule
\end{tabular}

\vspace{-15pt}
\label{table: different_hyps}
\end{table*}

\noindent\textbf{Effects of Hyperparameters.}
We study how certain hyperparameters' choices influence our method's performance. Key hyperparameters include: the number of rendered templates, the number of reference views selected for matching, the number of key frames selected for multi-frame optimization. Quantitative results are shown in \Tref{table: different_hyps}.
Increasing templates, reference views, and keyframes improves performance but slows inference, with diminishing returns at higher values.
As a balance between performance and inference speed, we use 960 total templates, 5 reference views, and 8 keyframes for optimization as our final method setup.

\subsection{Comparison with Robot Pose Estimation}
Our end-effector pose estimation algorithm ultimately serves as a \textbf{robot self-calibration algorithm}. Consequently, we compare our approach with learning-based camera-to-robot pose estimation algorithms, specifically RoboPose~\cite{labbe2021single}, CTRNet~\cite{lu2023markerless}, and RoboKeyGen~\cite{tian2024robokeygen}.
\Tref{table: compare_robopose} presents comparison results on the RealSense-Franka dataset. To obtain the end-effector pose from camera-to-robot pose estimation methods, we use the robot joint angles and forward kinematics to calculate the pose of the end effector in the robot space, and then transfer it to the camera space using the predicted camera-to-robot pose. The AUC threshold is raised to 2cm because all robot pose estimation methods yield near-zero results on the 1cm threshold.
% Even though the robot pose estimation methods have undergone robot-specific training on large amounts of data, our method outperforms all three baselines by a very large margin. 
Despite extensive robot-specific training, our method significantly outperforms all baselines.

\subsection{ Real-world high-precision targeting experiment}

Following~\cite{chen2023easyhec}, we conduct a high-precision targeting experiment to compare with marker-based hand-eye calibration method~\cite{tsai1989handeye}. Both our method and the marker-based method use the images and the robot arm's position to calculate the transformation from the camera coordinate system to the robot base coordinate system. We selected 20 positions for calibration, covering as much of the workspace as possible. For each specified number of views, we sample different views from the 20 positions and experiment 3 times to reduce variance. Additionally, the marker-based method requires a marker board to be attached to the end-effector, and our method uses Track-anything~\cite{yang2023track} for segmentation with a manually specified prompt for the first frame.

\begin{table}[h!]
\centering
\vspace{-4pt}
\caption{\textbf{Comparison with robot pose estimation methods on RealSense-Franka dataset.}}
\vspace{-5pt}
\resizebox{1.0\linewidth}{!}{

\begin{tabular}{cccc}
\toprule
Method           & Training-free & ADD@2cm & ADD-S@2cm \\
\midrule
\midrule
RoboPose\cite{labbe2021single}    & \No & 0.68 &  6.94  \\
CTRNet\cite{lu2023markerless}   & \No  & 5.46 & 22.57 \\
RoboKeyGen\cite{tian2024robokeygen} & \No & 9.35 & 29.99 \\
\colour \textbf{FEEPE(Ours)} & \Yes & \textbf{60.5} & \textbf{75.22} \\
\bottomrule
\end{tabular}
}
\vspace{-6pt}
\label{table: compare_robopose}
\end{table}

\begin{figure}[tbp]
\centering
\includegraphics[width=0.48\textwidth]{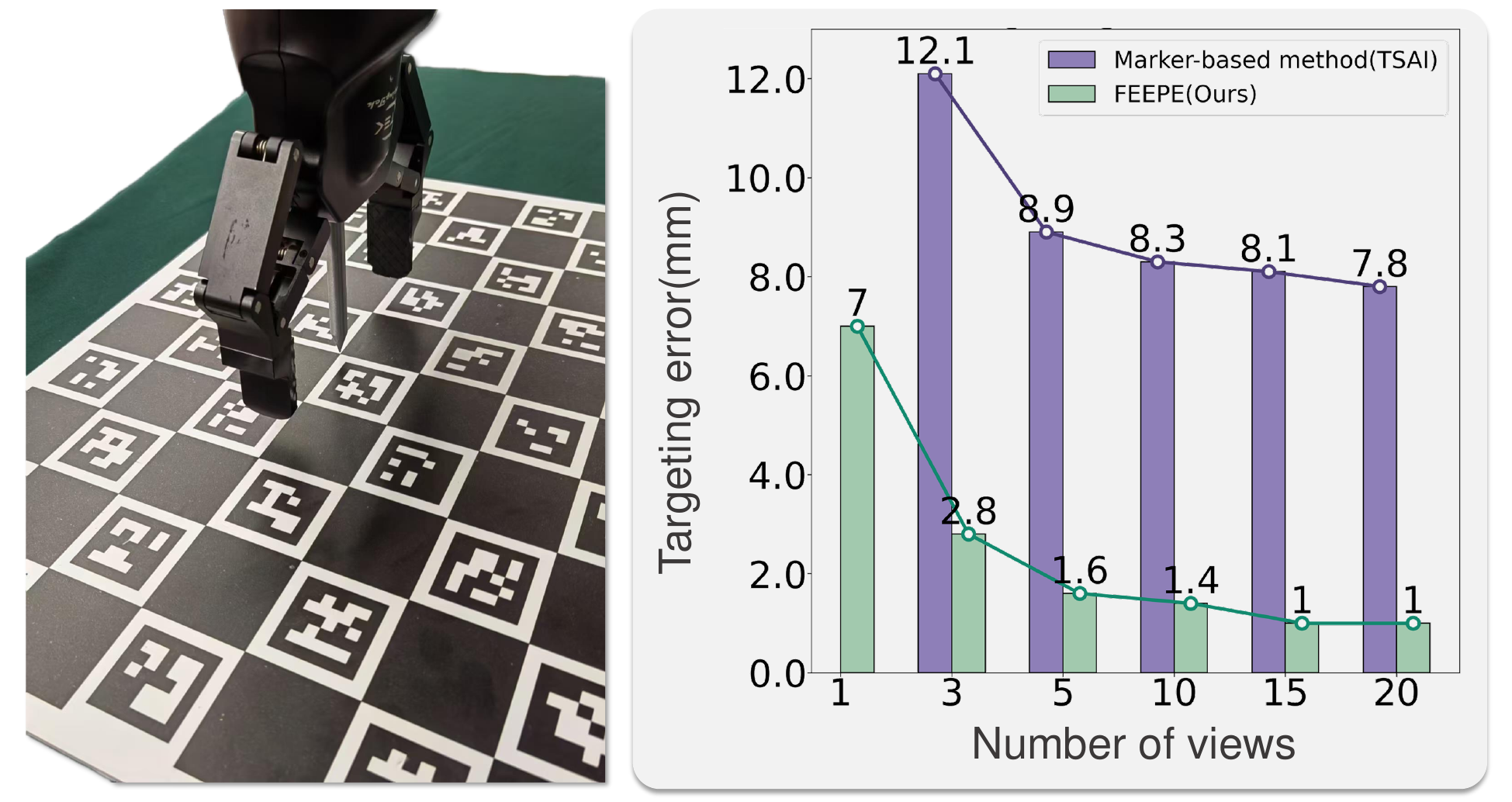}
\vspace{-18pt}
\caption{\textbf{Results of high-precision targeting experiments.}}
\vspace{-15pt}
\label{fig:targeting}
\end{figure}

As shown in ~\Fref{fig:targeting}, we detect the 5 different corners of the marker board using OpenCV~\cite{bradski2000opencv} and transform their positions to the robot base coordinate system via the calibration results from the process mentioned above. Then we use a pointer attached to the end-effector to tip these positions and manually measure the targeting error. The results in~\Fref{fig:targeting} demonstrate that our method consistently outperforms traditional marker-based calibration across all numbers of views. Moreover, our approach achieves an accuracy within 3mm using just three frames and reaches 1mm accuracy with 15 frames. Overall, compared to conventional hand-eye calibration methods, our method is not only marker-free and online but also significantly more precise.
Additionally, we also conduct real-world grasping experiments. Please refer to the supplementary materials for details.

\begin{table}[h!]
\begingroup
\centering
\vspace{4pt}
\caption{\textbf{Results of online grasping experiments.}}
\resizebox{1.0\linewidth}{!}{
\begin{tabular}{cccccccc}
\toprule
& \#1 & \#2 & \#3 & \#4 & \#5 & \#6 & Average \\ \midrule \midrule
MegaPose & 0.50 & 0.20 & 0.00 & 0.57 & 0.20 & 0.71 & 0.43 \\
SAM6D & 0.67 & 0.83 & 0.57 & 0.00 & 0.83 & 0.00 & 0.58 \\
FoundationPose & 0.57 & 0.83 & 0.71 & \textbf{1.00} & 0.50 & \textbf{1.00} & 0.75 \\
\colour \textbf{FEEPE(Ours)} & \textbf{1.00} & \textbf{1.00} & \textbf{1.00} & \textbf{1.00} & \textbf{1.00} & 0.83 & \textbf{0.97} \\
\bottomrule
\end{tabular}

}
\label{table: grasp_result}
\endgroup
\end{table}

\subsection{ Real-world online grasping experiment}
We validate the effectiveness of our method on the downstream task of robotic grasping~\cite{rgbgrasp}. Following~\cite{tian2023robot}, we use estimated end-effector poses to conduct grasping experiments without offline calibration. 30 different objects are categorized into 6 groups. We perform grasping experiments using GSNet~\cite{gsnet} three times with different camera poses for each group. Detailed object information and environment settings can be found on our website. As shown in \Tref{table: grasp_result}, our method consistently outperforms others in online grasping success rate, demonstrating high accuracy.

%%%%%%%%%%%%%%%%%%%%%%%%%%%%%%%%%%%%%%%%%%%%%%%%%%%%%%%%%%%%%%%%%%%%%%%%%%%%%%%%

\section{CONCLUSION}
\label{sec:conclusion}
In conclusion, we introduced FEEPE, a foundation feature-driven, online, training-free, and generalizable end-effector pose estimation method. Extensive experiments demonstrate that FEEPE outperforms learning-based and traditional methods, and provides superior flexibility and generalization, enabling effective online robot self-calibration.

\noindent\textbf{Limitations.} Our approach relies on prior knowledge of the end-effector's CAD model and real-time state feedback, limiting its flexible application across diverse scenarios. Additionally, our method requires prior segmentation of the end-effector, necessitating an extra segmentation module. Future research could explore mask-free settings.

% \begin{table}[h!]
% \centering
% \input{Tables/grasp_result}
% \caption{grasp result.}
% \label{table: grasp_result}
% \end{table}

%%%%%%%%%%%%%%%%%%%%%%%%%%%%%%%%%%%%%%%%%%%%%%%%%%%%%%%%%%%%%%%%%%%%%%%%%%%%%%%%

\section{ACKNOWLEDGMENT}
\label{sec:acknowledgment}
This work is supported by the National Youth Talent Support Program (8200800081) and National Natural Science Foundation of China (No. 62376006). We would like to thank Yang Tian from PKU for the valuable technical support.

\clearpage
{
\bibliographystyle{IEEEtran}
\bibliography{IEEEabrv, main}
}

\end{document}